\documentclass[conference]{IEEEtran}
\IEEEoverridecommandlockouts
\usepackage{cite}

\usepackage[ruled,vlined]{algorithm2e}
\usepackage{amssymb}
\usepackage{latexsym}
\usepackage{framed}
\usepackage{multirow}
\usepackage{amsmath,amssymb,amsfonts}
\usepackage{algorithmic}
\usepackage{graphicx}
\usepackage{adjustbox}
\usepackage{textcomp}
\usepackage{xcolor}
\usepackage{lipsum}
\usepackage{booktabs} 

\usepackage{hyperref}
\hypersetup{
    colorlinks=true,
    linkcolor=black,
    filecolor=magenta,      
    urlcolor=black,
    citecolor=black,
}

\usepackage{url}
\def\BibTeX{{\rm B\kern-.05em{\sc i\kern-.025em b}\kern-.08em
    T\kern-.1667em\lower.7ex\hbox{E}\kern-.125emX}}

\makeatletter

\def\ps@IEEEtitlepagestyle{%
    \def\@oddfoot{\mycopyrightnotice}%
    \def\@evenfoot{}%
}
\def\mycopyrightnotice{%
    {\footnotesize  978-1-6654-3405-8/21/\$31.00 \textcopyright2021 IEEE\hfill}
    \gdef\mycopyrightnotice{}
}



\title{An enhanced method of initial cluster center selection for K-means algorithm}

\begin{document}
\makeatletter
\newcommand{\linebreakand}{%
    \end{@IEEEauthorhalign}
    \hfill\mbox{}\par
    \mbox{}\hfill\begin{@IEEEauthorhalign}
    }
\makeatother

\author{
\IEEEauthorblockN{1\textsuperscript{st} Zillur Rahman}
\IEEEauthorblockA{\textit{Department of Electrical and Electronic Engineering} \\
\textit{Chittagong University of Engineering and Technology}\\
Chittagong, Bangladesh \\
u1502056@student.cuet.ac.bd}
\and
\IEEEauthorblockN{2\textsuperscript{nd} Md. Sabir Hossain}
\IEEEauthorblockA{\textit{Department of Computer Science and Engineering} \\
\textit{Chittagong University of Engineering and Technology}\\
Chittagong, Bangladesh \\
sabir.cse@cuet.ac.bd}
\and
\linebreakand
\IEEEauthorblockN{3\textsuperscript{rd} Mohammad Hasan}
\IEEEauthorblockA{\textit{Department of Computer Science and Engineering} \\
\textit{Chittagong University of Engineering and Technology}\\
Chittagong, Bangladesh \\
hasancse.cuet13@gmail.com}
\and
\IEEEauthorblockN{4\textsuperscript{th} Ahmed Imteaj}
\IEEEauthorblockA{\textit{Department of Computer Science and Engineering} \\
\textit{Chittagong University of Engineering and Technology}\\
Chittagong, Bangladesh \\
aimteaj@gmail.com}

}

\maketitle

\begin{abstract}
Clustering is one of the widely used techniques to find out patterns from a dataset that can be applied in different applications or analyses. K-means, the most popular and simple clustering algorithm, might get trapped into local minima if not properly initialized and the initialization of this algorithm is done randomly. In this paper, we propose a novel approach to improve initial cluster selection for K-means algorithm. This algorithm is based on the fact that the initial centroids must be well separated from each other since the final clusters are separated groups in feature space. The Convex Hull algorithm facilitates the computing of the first two centroids and the remaining ones are selected according to the distance from previously selected centers. To ensure the selection of one center per cluster, we use the nearest neighbor technique. To check the robustness of our proposed algorithm, we consider several real-world datasets. We obtained only 7.33\%, 7.90\%, and 0\% clustering error in Iris, Letter, and Ruspini data respectively which proves better performance than other existing systems. The results indicate that our proposed method outperforms the conventional K means approach by accelerating the computation when the number of clusters is greater than 2.
\end{abstract}


\section{Introduction}
Clustering is a technique of grouping data in accordance to the features such that data within same clusters have maximum similarities and data of different clusters have minimum similarities. This technique is used in tons of applications like data analysis, image segmentation, vector quantization, and pattern recognition. There are several different clustering algorithms such as hierarchical clustering, partitioning based, grid-based, and density-based clustering algorithm \cite{rai2010survey}. Among them, partitioning based algorithms are widely used.\\
K-means is a famous partitioning based clustering algorithm that groups the data samples into K number of disjointed clusters \cite{macqueen1967some}. If the dataset contains N number of samples such as $x_1, x_2....x_N$ and there are K clusters such as $C_1, C_2...C_K$, then the cost function used in K-means is defined by (\ref{eqn1}) where $x_j$ is a data sample of cluster $C_i$ and $c_i$ is the centroid of that cluster.
\begin{equation}
J = \sum_{i=1}^{K}\sum_{x_j\epsilon C_i}{d(x_j,c_i)}
\label{eqn1}
\end{equation}
The centroid of each cluster is the center of mass of that cluster computed using (\ref{eqn2}) where $|C_i|$ is the number of samples in cluster $C_i$; $0<|C_i|<N$
\begin{equation}
c_i = \frac{1}{|C_i|}\sum_{x_j\epsilon C_i}{x_j}
\label{eqn2}
\end{equation}

The detail of the simple K-means algorithm is as follows:
\begin{enumerate}
\item{Take K and dataset as input and repeat process 2-5 for a certain number of iterations.}
\item{Randomly select K samples from the dataset as the initial cluster centers and repeat process 3-5 until convergence.}
\item{Calculate the Euclidean distance between each data samples and each centroid.}
\item{Assign each data sample to its nearest cluster.}
\item{Calculate the centroid of each cluster using (\ref{eqn2})}
\item{Compute the error using (\ref{eqn1}) and take those initial centroids corresponding to the minimum error.}
\end{enumerate}

From the above algorithm, it is clearly understood that the error depends on the initial random centroids. Each random selection results in a different amount of error and finding the best solution might take a lot of iterations and computational time as well. This is the biggest limitation of the k-means algorithm.\\
In this paper, we propose a simple and yet efficient approach to select the initial cluster centers which result in very fast convergence along with excellent clustering performance in four real-world datasets.\\
The rest of the paper is organized as follows. The \autoref{sec2} briefly describes the existing related systems, and \autoref{sec3} presents our proposed algorithm in detail. The results obtained from the four experiments along with dataset descriptions are in \autoref{sec4}. The paper ends with the conclusion in \autoref{sec5}.
\section{Related Works}\label{sec2}
Several systems have been developed to find better initialization for the K-means clustering algorithm. A novel approach was presented to select the initial centroids of the clusters in \cite{bradley1998refining}. A set of sub-samples are created from the entire dataset, and K-means is applied to each sub-sample. In that way, (K x J) number of centroids will be found for the J number of sub-samples. Then, K-means is again applied to these centroids using K centroids of each sub-sample as initialization. The centroids that give minimum error will be used as the initial centroids for the whole dataset. This method is proved to be useful for large datasets.\\
A fast global k-means algorithm was proposed in \cite{likas2003global} where a global search technique was developed which adds one cluster center at a time. The process is repeated until the K number of centroids are found and principle component analysis(PCA) is used for dimensionality reduction of different real-world datasets.\\
A very simple initial centroid selection algorithm was developed in \cite{arthur2006k}. The first centroid is selected randomly from all the samples. The second centroid is the sample having maximum distance from its nearest centroid. This process continues until the number of initial centroids reaches K.\\
Another method for selecting initial centroids was proposed in \cite{yuan2004new}. All the Euclidean distances between the samples are computed and the two samples with the shortest distance are selected. Then these two points are deleted from the main sample set along with their n nearest neighbor. In that way, a new set will be created. By repeating this process for K times, K sets are created. The mean of each set is used as an initial centroid for the K-means algorithm.\\
In \cite{khan2004cluster}, the authors proposed Cluster Center Initialization Algorithm (CCIA) to select initial centroids. In this system first, all the attributes are normalized and one attribute is selected. The mean, standard deviation, and percentile value from the normal curve are used to find K number of centers for that attribute. Then, K-means is applied for that attribute only and each sample is labeled by a cluster number. By repeating this for the remaining attributes, a pattern is created for each sample. The unique patterns are then identified and samples corresponding to such each pattern are assigned to a single cluster.\\
Another approach was presented for initial centroid selection where the whole dataset is considered as a single cell and the axis with the highest variance is selected as the principal axis in \cite{deelers2007enhancing}. The cell is divided into two with help of a cutting plane and error is calculated before and after the division. In that way, cells are divided until the number of cells is the same as the number of clusters, and the mean of each cell is then used as the initial centroids for the K-means algorithm.\\
A novel approach for selecting the number of clusters was proposed in \cite{vzalik2008efficient}. The author developed a new cost function as well as a new function for assigning each sample to a cluster. The simple K-means is applied for different K and the value respective to the global minimum of the cost function is the actual number of cluster. Each sample is included in a cluster if the corresponding function value for that cluster is minimum.\\
The authors in \cite{huang2012improved} presented another technique for initial centroid selection where all the distances between samples are calculated first. The mean distance and density of each sample are computed using the developed density function. The sample having maximum density is selected as the first centroid and some neighbors of this sample are deleted from the stored density array. This process is repeated K times.\\
A simple technique based on the weighted average of the attributes was developed in \cite{mahmud2012improvement}. The attribute values are weighted averaged for each sample and sorted. The sorted list is then divided into K set and the mean of each set will be the initial centroids for simple K-means algorithm.\\
The authors in \cite{motwani2019study} proposed a new initial centroids selection method. First, all the distances between samples are calculated. Then the sums of distances for all points are computed and sorted using their sum of distances from maximum to minimum. The point, having the maximum sum of distances is the first centroid. Then next N/K samples are discarded and the first sample after that is selected as the second initial centroid. This process will be for K times.\\
Another solution for initial centroid selection was presented in \cite{erisoglu2011new}. This algorithm is based on the farthest samples. The mean of the dataset is calculated and the farthest sample from the mean is selected as the first initial centroid. The sample having maximum distance from the previously selected centroids is the second centroid and so on.
\section{Methodology}\label{sec3}
This section explains the proposed algorithm to compute the initial centroids for the K-means. To illustrate the procedures, we generated a small dataset using random samples namely synthetic dataset 1 that consists of two attributes and five clusters. There is total 35 samples, and we named the attributes variable1 and variable2. The dataset is shown in Fig. \ref{fig01}.
\begin{figure}[b]
\centerline{\includegraphics[width=7cm,height=5.0cm]{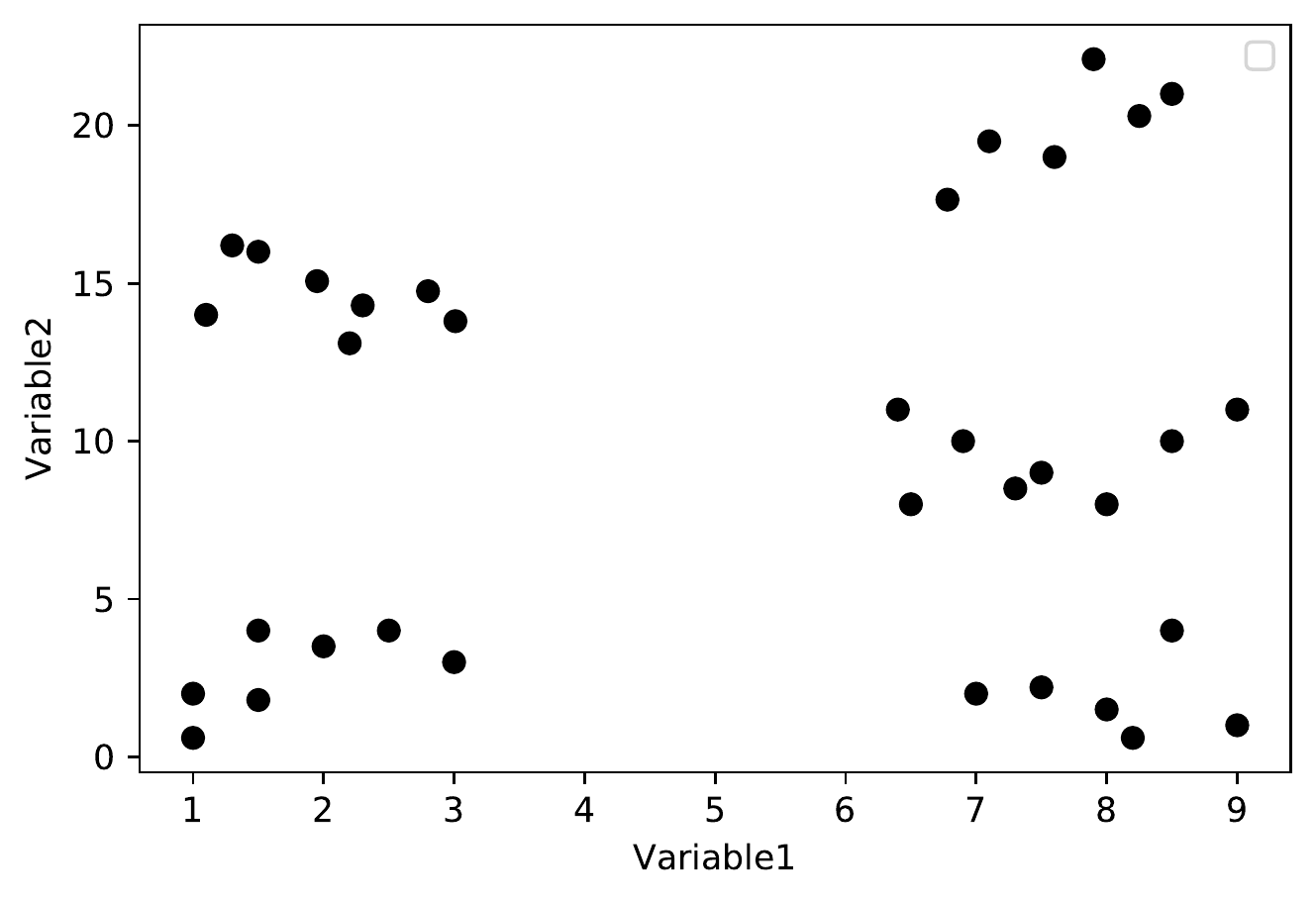}}
\caption{Synthetic dataset 1}
\label{fig01}
\end{figure}
 
The first two initial centroids certainly are far away from each other. We can find the two farthest samples by calculating distance from all samples to all samples. Unfortunately, this is a very time consuming process. However, we can reduce this computation if we consider only the outer samples of the dataset because outer points contain the first two initial centroids. To get the outer samples, we used the Convex Hull algorithm \cite{barber1996quickhull}, which provides the smallest convex containing all the samples in the data. The polygon contains the outer samples which is presented in Fig. \ref{fig02}.\\ 
\begin{figure}[t]
\centerline{\includegraphics[width=7cm,height=5.0cm]{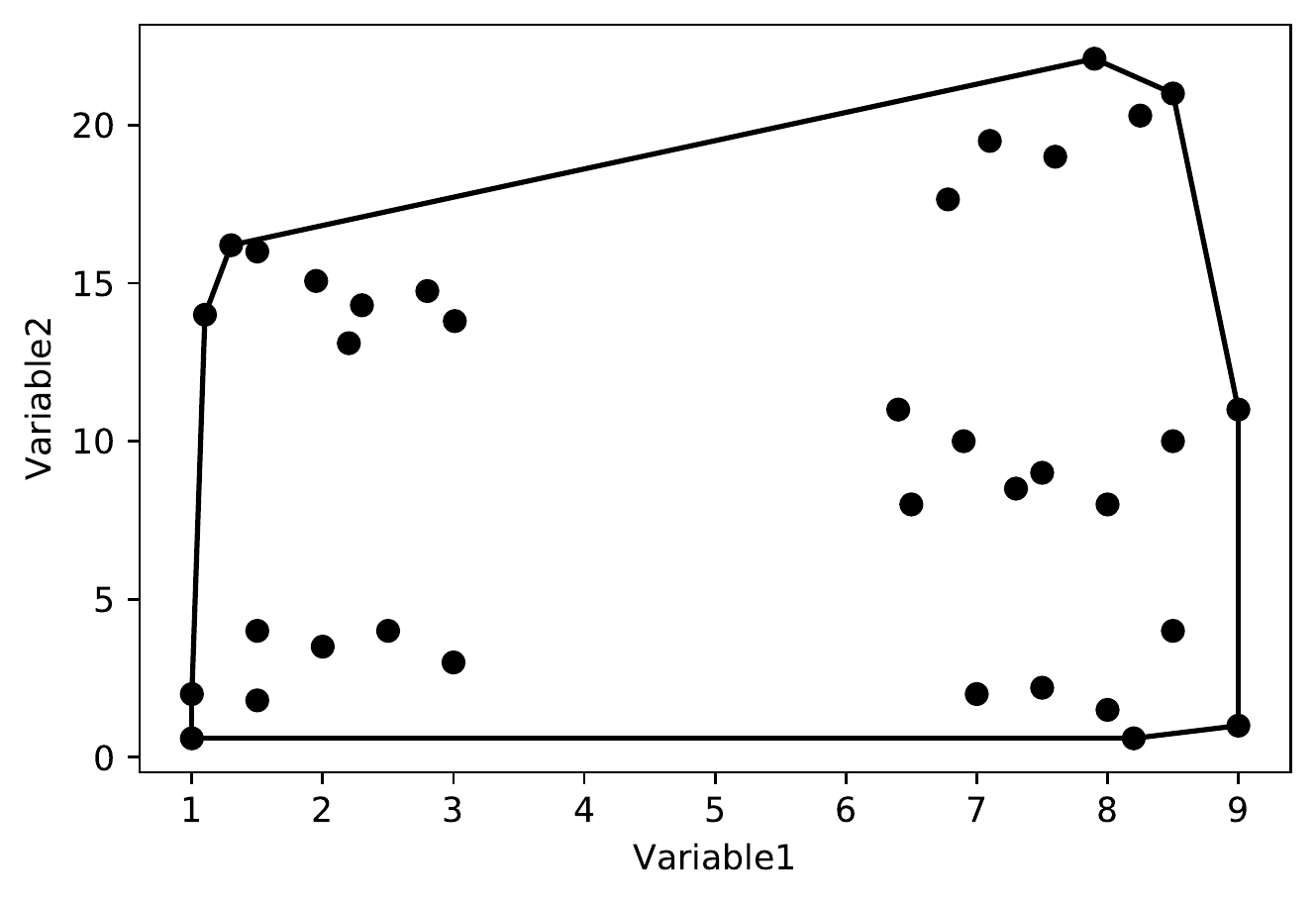}}
\caption{The convex hull algorithm }
\label{fig02}
\end{figure}
Now, for finding the two farthest samples $c_1$ and $c_2$, we computed all the possible distances between only the outer points and selected the two having maximum distance which is illustrated in Fig. \ref{fig03}(a). These two samples will be our first two initial centroids. The third initial centroid $c_3$ will be the sample $x_j$ having maximum distance from the previous two centroids, computed using (\ref{eqn3}) and is shown in Fig. \ref{fig03}(b).
\begin{equation}
d_3 = max( d(c_1, x_j)+ d(c_2, x_j) ) 
\label{eqn3}
\end{equation}
\begin{figure}[h]
\centerline{\includegraphics[width=9cm,height=7cm]{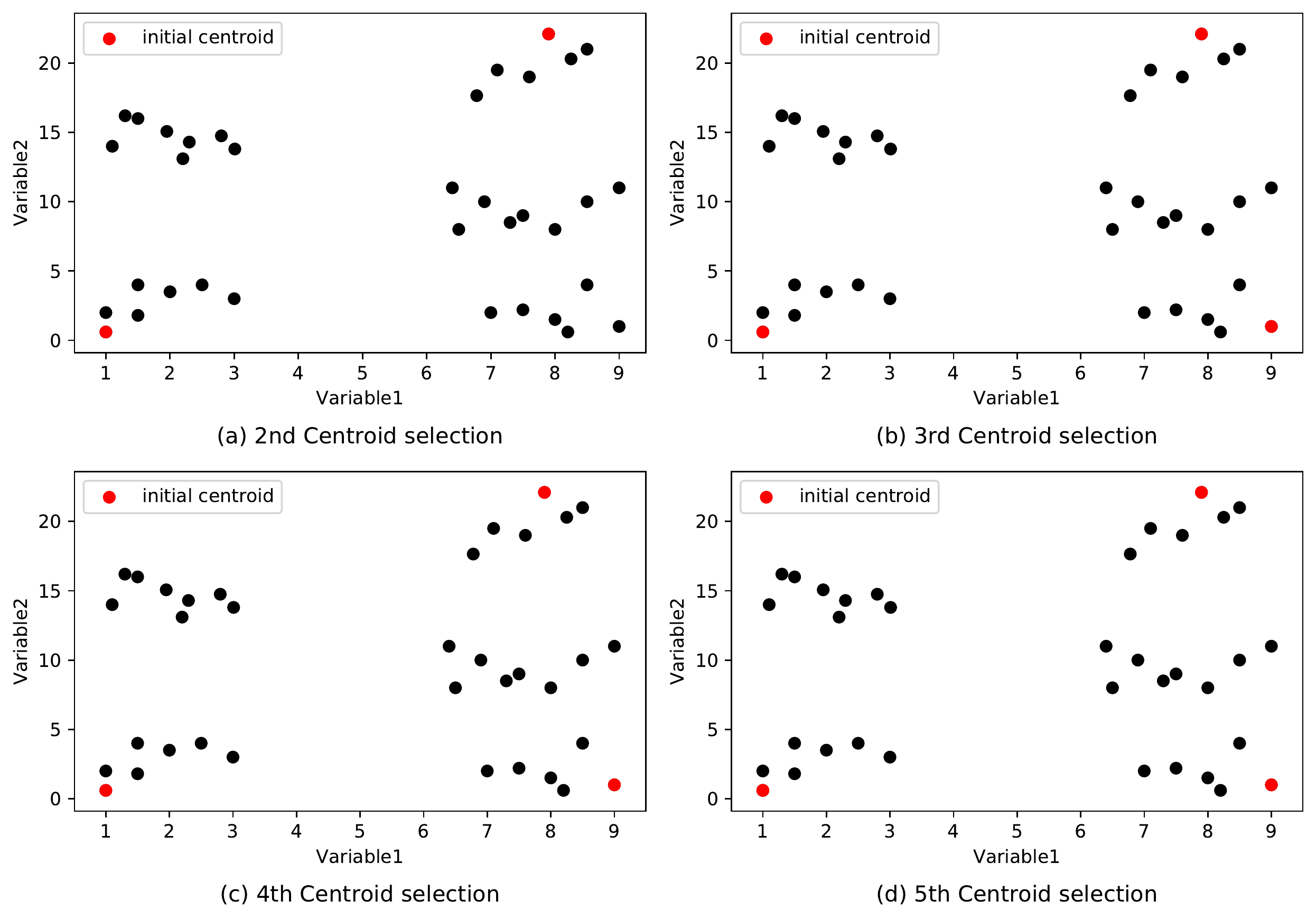}}
\caption{The other initial centroids}
\label{fig03}
\end{figure}

\begin{figure}[h]
\centerline{\includegraphics[width=8cm,height=5.0cm]{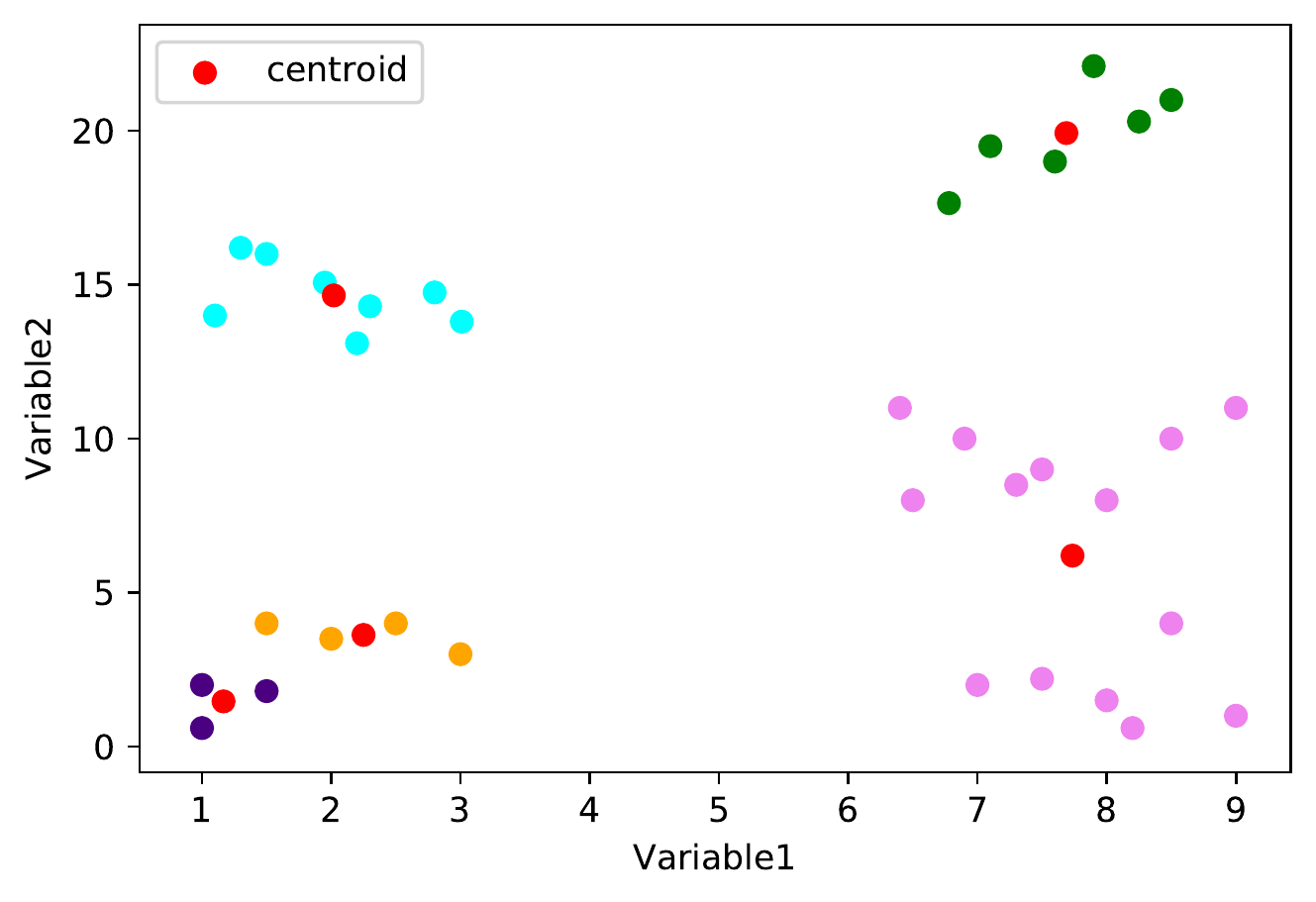}}
\caption{The final clustering result of synthetic dataset 1}
\label{fig04}
\end{figure}
While calculating the fourth initial centroids in this way like, we encounter a problem. The fourth one turns out to be the first centroid itself which is shown in Fig. \ref{fig03}(c) since it has the maximum sum of distances from the all the previous three. If we do not use the previous centroids during the new centroid selection, then a close one in the same cluster of one of the previous centroids might get selected. The same is true for the fifth centroid, shown in Fig. \ref{fig03}(d). Since the initial centroids are not in the right place, the clustering result shows poor performance which is evidenced from Fig. \ref{fig04}.

To eliminate the chance of selecting the two centroids from the same cluster, we discarded M nearest samples of the previously selected centers during new centroid computation. Empirical studies showed M is in the range of m to m/3 depending on data distribution, where m is $N/K$. As a result, the probability of choosing a centroid from a new cluster increases.
The fourth and remaining centroids are selected based on (\ref{eqn5}) and is depicted in Fig. \ref{fig05}. We can now select the necessary number of initial centroids and use them in simple K-means clustering algorithm.
\begin{equation}
d_k = max( \sum_{i=1}^{k-1}{d(c_i, x_j)} )
\label{eqn5}
\end{equation}
The final clustering result presented in Fig. \ref{fig06} shows error-free performance. The steps involved in our proposed algorithm are provided in algorithm \autoref{alg1}.

\begin{figure}[h]
\centerline{\includegraphics[width=9cm,height=7cm]{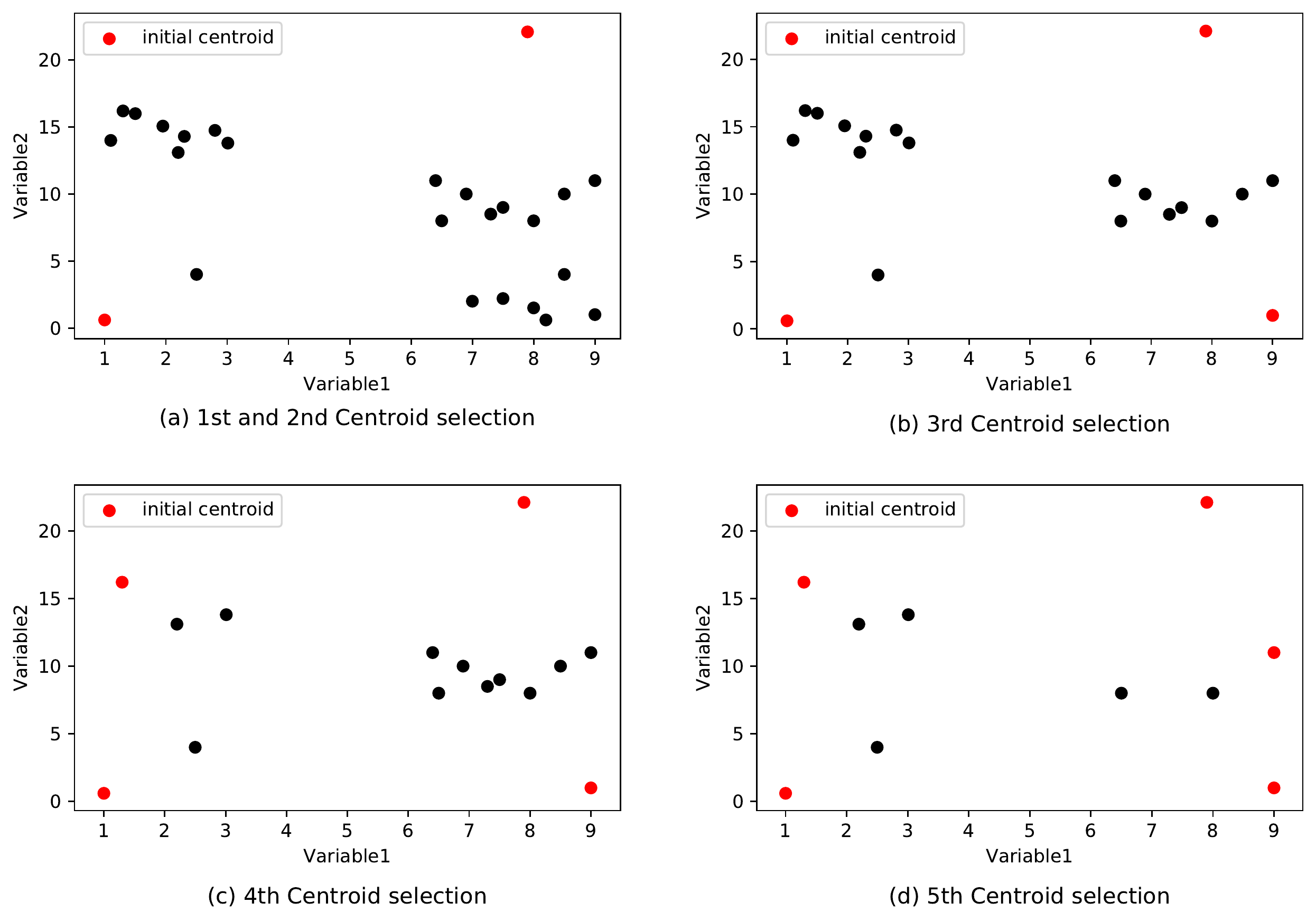}}
\caption{The other initial centroids}
\label{fig05}
\end{figure}

\begin{figure}[h]
\centerline{\includegraphics[width=8cm,height=5.0cm]{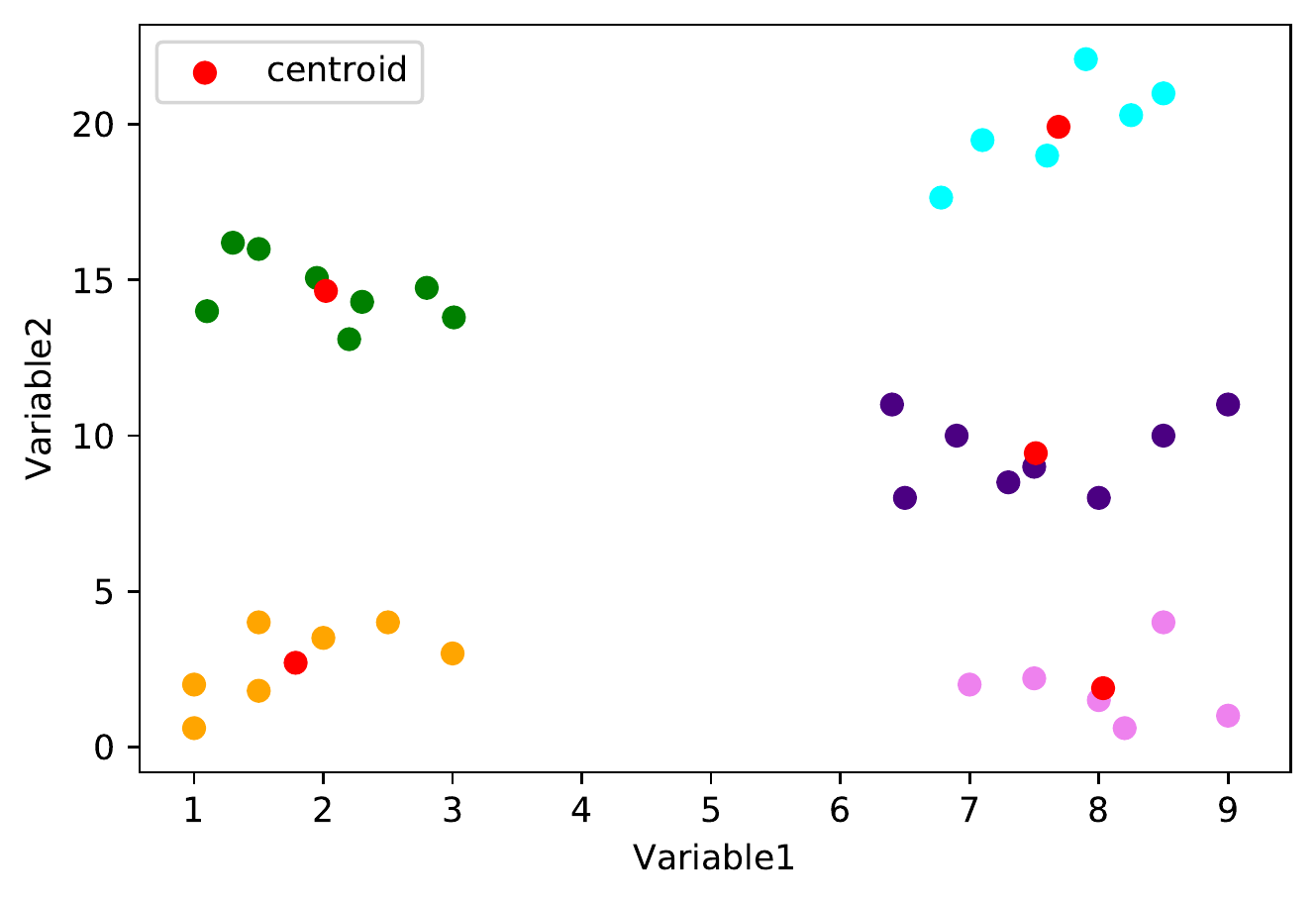}}
\caption{The final clustering results}
\label{fig06}
\end{figure}

\begin{algorithm}[!h]
\caption{Proposed Algorithm}
\hspace*{\algorithmicindent} \textbf{Input:} Dataset S, number of clusters K, and number of discarded samples M\\
\hspace*{\algorithmicindent} \textbf{Output:} The resulted clusters
\begin{algorithmic}[1]
\STATE Compute convex hull of the dataset\;
\STATE Calculate Euclidean Distance between outer points\;
\STATE Take two points $c_1$ and $c_2$ with maximum distance as the first two initial centroids\;
\STATE Discard M nearest neighbors of the two selected centroids\;
\FOR{i=1:K-2}
\STATE Find farthest sample $c_{i+2}$ from previously selected centroids using (\ref{eqn5})\;
\STATE Discard M nearest neighbors of the previously selected centroids\;
\ENDFOR
\STATE Use these centroids as the initial cluster centers for K-means algorithm\;
\end{algorithmic}
\label{alg1}
\end{algorithm}

\section{Experimental Results and Discussion}\label{sec4}
To evaluate our proposed model, we implemented it in Python 3.74 and tested its performance in one synthetic and four real-world datasets including Iris data, Wine data, and Letter image recognition from the UCI machine learning repository \cite{Dua:2019} and Ruspini data \cite{ruspini1970numerical}. These datasets contain labels for each sample, and we used those for evaluation of our model.\\
The synthetic dataset 2 consists of 300 samples and 6 clusters of 0.75 standard deviation. The clustering result is presented in Fig. \ref{fig07} which shows an excellent performance with only 2 misclassified samples.\\
The Iris dataset\cite{Dua:2019}, which was taken from the UCI machine learning repository, is one of the most widely used datasets for testing the clustering algorithm. This dataset has three classes, 50 samples per class, and a total of 150 samples. Each class represents one kind of Iris flowers such as Iris setosa, Iris versicolor, and Iris virginica. Each sample has four attributes sepal length, sepal width, petal length, and petal width.\\
The Wine dataset\cite{Dua:2019} is another popular dataset taken from the same UCI repository that describes the constituents found in three types of wine. There is a total of 178 instances with 59, 71, and 48 instances in class I, II, and III respectively. Each type of wine has 13 constituents. Since not all the attributes are important, we used PCA for dimensionality reduction which results in fast convergence with satisfactory performance.\\
The letter image recognition data\cite{Dua:2019} from the UCI repository has a total of 26 classes and 20,000 samples. The objective of this dataset is to identify each one of the English alphabet and each class represents a single alphabet from A-Z. The image of each sample was collected from 20 different fonts and by randomly distorting those images 20,000 samples were created. Each sample has 16 attributes scaled to fit into a range of 0 to 15. In this experiment, we only used 789 samples of letter A and 805 samples of letter D to ease the comparison with other existing systems and PCA as well.\\
The Ruspini dataset\cite{ruspini1970numerical} consists of a total of 75 instances and each instance has two attributes. The dataset has four classes: 23, 20, 17, and 15 samples in class I, II, III, and IV respectively.\\

\subsection{Evaluation metrics}
The main metric we used for the evaluation process is the Error which is defined as the percentage of misclassified samples refers to (\ref{eqn6}).
\begin{equation}
Error (\%) = \frac{Number\ of\ misclassified\ samples}{Total\ samples}\ X\ 100
\label{eqn6}
\end{equation}

To measure the degree of closeness between the initial centroids generated by our proposed algorithm or random initialization and the actual cluster center, we used the Cluster Center Proximity Index (CCPI) like \cite{khan2004cluster} defined by (\ref{eqn7}).
\begin{equation}
CCPI = \frac{1}{K*n}\sum_{i=1}^{K}\sum_{j=1}^{n}\frac{|A_{ij}- C_{ij}|}{|A_{ij}|}
\label{eqn7}
\end{equation}
Here, K is the number of clusters, n is the number of variables, A is the actual cluster center and C is the cluster center generated by our proposed algorithm.

The Rand index \cite{rand1971objective} has been used to measure the degree of similarity between two data clustering. Given a dataset $X = \{x_1,...,x_n\}$, suppose $A = \{a_1,..a_k\}$ and $B = \{b_1,..,b_k\}$ represent partitions of the objects with our proposed algorithm and actual cluster memberships, respectively. If \{xi,xj\} is a random pair of elements:\\
a: the number of pairs in X that are in the same subset in both A and B\\
b: the number of pairs in X that are in different subsets in A but in different subsets in B\\
c: the number of pairs in X that are in the same subset in A but in different subsets in B\\
d: the number of pairs in X that are in different subsets in both A and B\\
The rand score is calculated using (\ref{eqn8}). The value is in between 0 and 1 and the closer it to 1, the higher the similarity between actual and proposed clustering.
\begin{equation}
Rand\ score = \frac{a+d}{a+b+c+d}
\label{eqn8}
\end{equation}

\subsection{Results}
The clustering result of the synthetic dataset 2 is illustrated in Fig. \ref{fig07} which shows an excellent performance with only 2 misclassified samples.  
\begin{figure}[h]
\centerline{\includegraphics[width=7.5cm,height=5.0cm]{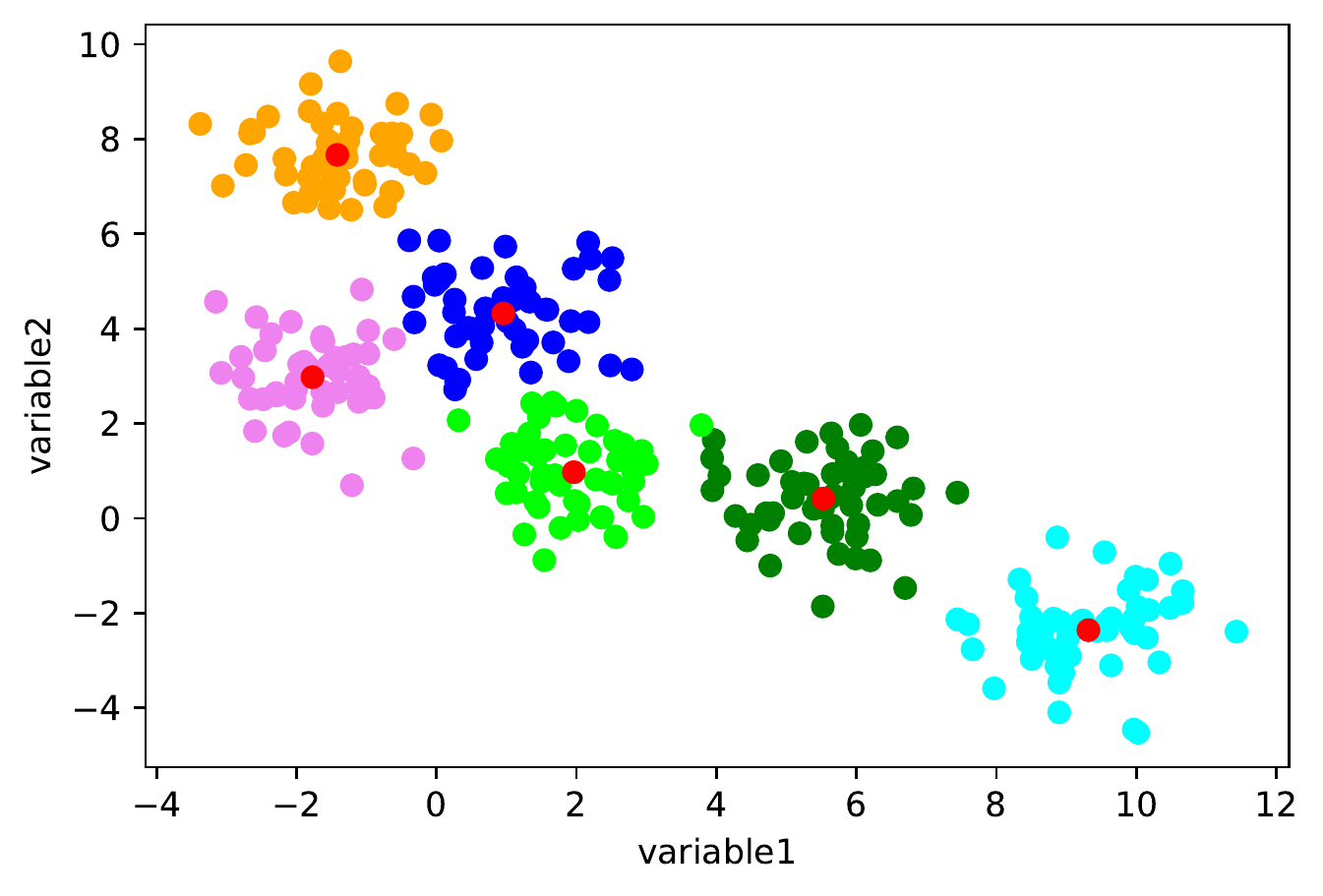}}
\caption{The synthetic dataset 2}
\label{fig07}
\end{figure}

The entire clustering process for the Iris dataset is shown step by step in Fig. \ref{fig09} to Fig. \ref{fig10}. At first, convex hull was computed which is shown in Fig. \ref{fig09}. Then, three centroids were calculated which are shown in Fig. \ref{fig10}. The final clustering result is illustrated in Fig. \ref{fig11}. From this figure, we can clearly see that there are only 11 misclassified samples among 150 samples.
\begin{figure}[h]
\centerline{\includegraphics[width=7cm,height=5.0cm]{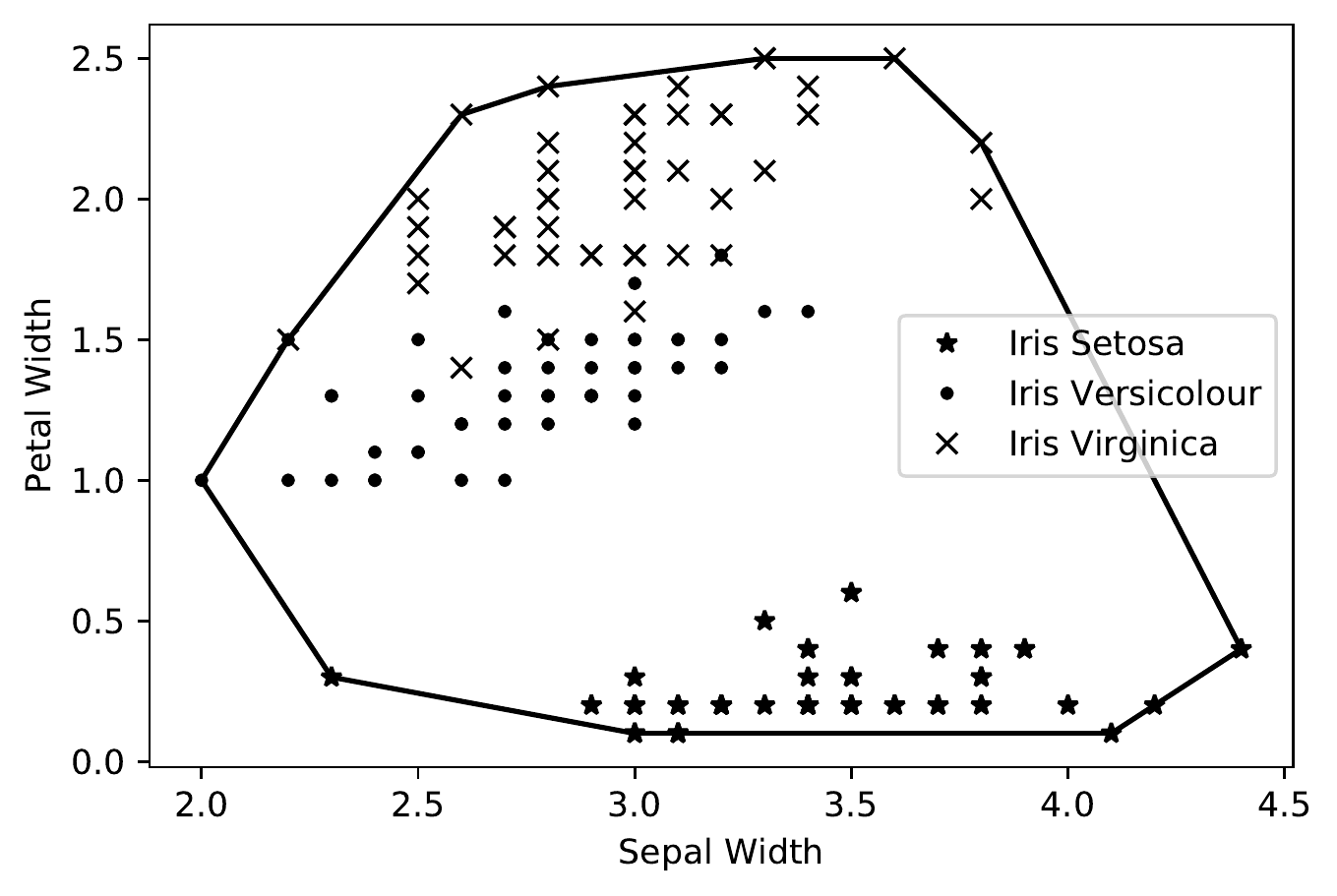}}
\caption{The convex hull computation}
\label{fig09}
\end{figure}
\begin{figure}[h]
\centerline{\includegraphics[width=9cm,height=5.0cm]{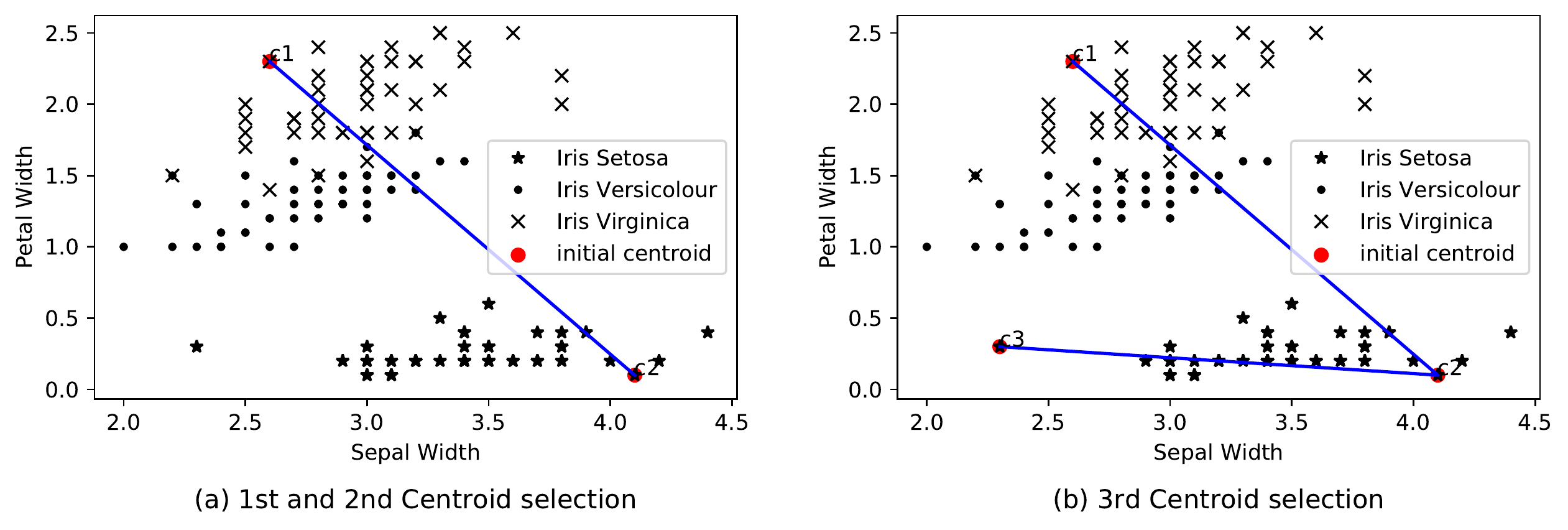}}
\caption{The initial centroids selection}
\label{fig10}
\end{figure}
\begin{figure}[h]
\centerline{\includegraphics[width=8cm,height=4.5cm]{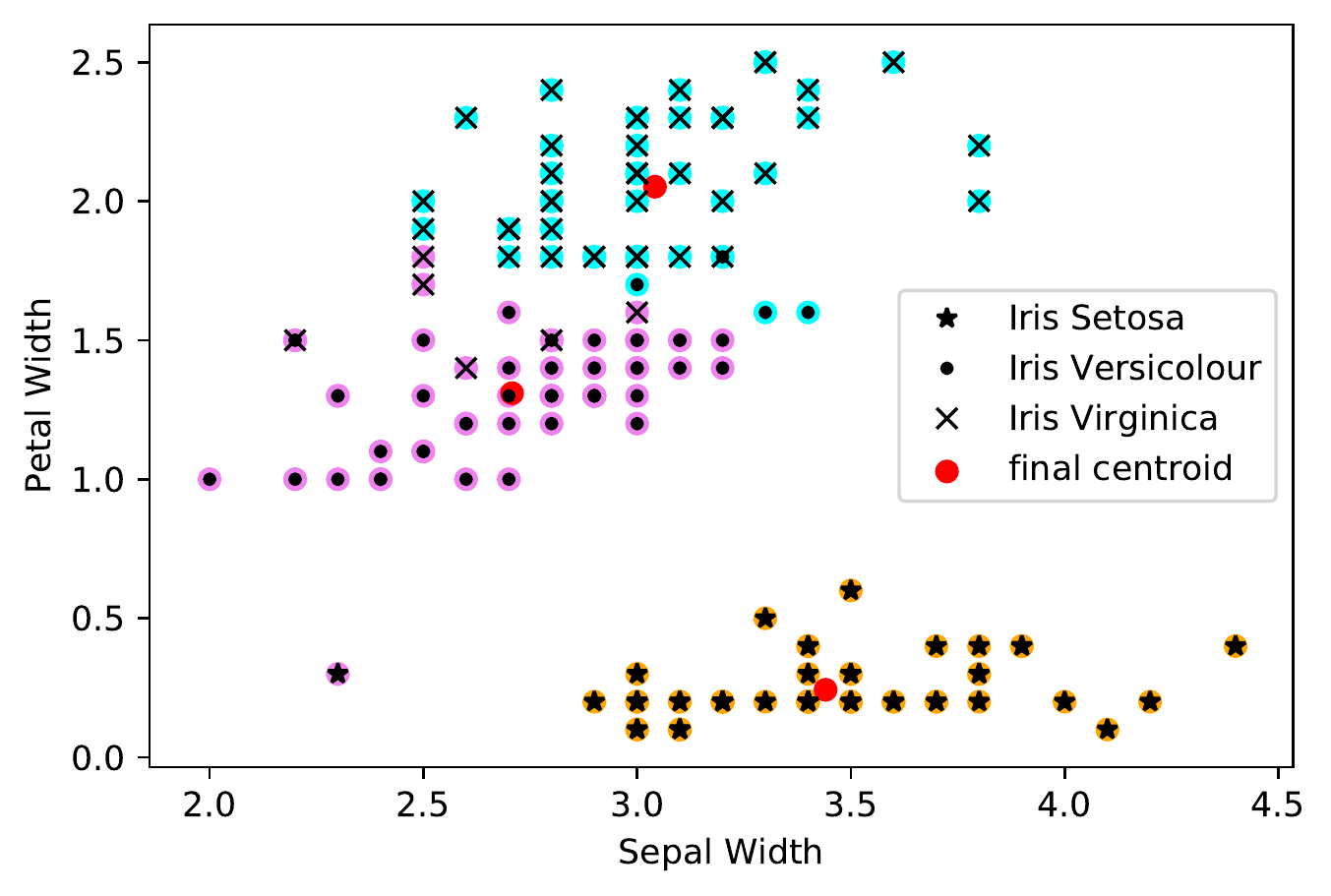}}
\caption{The final result of the Iris dataset}
\label{fig11}
\end{figure}

\begin{table}[!h]
\caption{CCPI and Computation time for all the datasets}
\begin{center}
\begin{tabular}{|c|c|c|c|c|}\hline

& \multicolumn{2}{c}{\textbf{Computation time(ms)}} & \multicolumn{2}{|c}{\textbf{CCPI}}\\\hline
Dataset & K-means & Proposed & K-means & Proposed \\\hline
Synthetic 2 & 1426.5  & 608.5 & 1.7528& 0.1957 \\\hline
Iris & 419.4 & 323 & 0.8417 & 0.3323\\\hline
Wine & 523.1 & 439 & 1.6134 & 0.7869\\\hline
Letter & 9170 & 13700 & 0.3527 & 0.4327\\\hline
Ruspini & 250 & 110.7 & 0.2584 & 0.0713\\\hline

\end{tabular}
\label{tab2}
\end{center}
\end{table}
The computation time and CCPI of each dataset are reported in \autoref{tab2}. The proposed algorithm is compared with the basic K-means algorithm which was initialized randomly and executed 20 times, then we took the mean results. The computation time values presented in \autoref{tab2} shows, when there is a fewer number of clusters such as 2, simple K-means works faster than our proposed algorithm. For example, the computation time for the Letter data is much lower for K-means than the proposed one. It's because 2 initial random centers have more chance to be close to actual ones which leads to fewer iteration to converge and finally results in less computation time. But when the number of cluster increases, the centroids generated by our proposed algorithm are highly likely to be close to the actual centers than centroids generated by random initialization and results in faster convergence which is evident from the computation time of rest of the datasets. On the other hand, for random initialization with many clusters, it would be normal that one cluster has more than one random center which leads to slower convergence. The same logic is true for CCPI values. When cluster number is 2, due to the high probability of being one random center per cluster, the random centers are close to the actual ones and CCPI is low for K-means. In contrary, our proposed algorithm shows excellent performance when the number of cluster increases since in that case the initial centers are within the corresponding clusters, near to actual one and CCPI values are low.\\
\autoref{tab3} shows the clustering error comparison of our system with other existing systems. Our proposed system performed better than others in Iris, Letter, and Ruspini data. The system in \cite{erisoglu2011new} shows good results in Letter and Ruspini data but unlike our system, the process is more complex since it requires the computation of principal axis and secondary axis first and then more distance calculations which leads to higher computation time. 
\begin{table}[h]
\caption{Comparison with other existing systems using Clustering error(\%)}
\begin{center}
\begin{tabular}{|c|c|c|c|c|c|}\hline
\textbf{System}  & \textbf{Iris} & \textbf{Wine} & \textbf{Letter} & \textbf{Ruspini}\\\hline
\cite{yuan2004new} & 11.33 & 31.5 & - & -\\\hline
\cite{khan2004cluster} & 11.33 & 5.05 & 8.60 & 4.0 \\\hline
\cite{deelers2007enhancing} & 11.0 & 5.05 & 8.60 & 4.0 \\\hline
\cite{vzalik2008efficient} & - & \textbf{2.25} & - & - \\\hline
\cite{huang2012improved} & 8.0 & 21.44 & - & - \\\hline
\cite{erisoglu2011new} & 10.70 & 3.40 & \textbf{7.90} & \textbf{0.0} \\\hline
\cite{han2010improved} &8.0 & 27.53 & - & -\\\hline
\cite{wang2011improved} & 18.5  & 17 & -& -\\\hline
Proposed & \textbf{7.33} & 3.37 & \textbf{7.90} & \textbf{0.0}
\\\hline
\end{tabular}
\label{tab3}
\end{center}
\end{table}

For further analysis, we report the rand index for all real world datasets in \autoref{tab4}. Our proposed system performs almost same for three dataset except Iris where our system shows clear superiority.

\begin{table}[h]
\caption{Rand index}
\begin{center}
\begin{tabular}{|c|c|c|c|c|c|}\hline
\textbf{System}  & \textbf{Iris} & \textbf{Wine} & \textbf{Letter} & \textbf{Ruspini}\\\hline
\cite{erisoglu2011new} & 0.8797 & 0.9543 & 0.8543 & 1.00\\ \hline
Proposed & 0.9104 & 0.9533 & 0.8532 & 1.00
\\\hline
\end{tabular}
\label{tab4}
\end{center}
\end{table}

\section{Conclusion}\label{sec5}
In this paper, we propose an enhanced K-means algorithm for iterative clustering with a better initialization technique. The technique is based on the fact that the cluster centers will be far away from each other. Since finding the two farthest samples by computing all the possible distances between samples is computationally expensive, we exploited the Convex Hull algorithm which provides us the set of outer samples. From that set, we have our first two initial centroids. The rest will be the samples farthest from the previously selected centers. Then, these centroids will be used as the initialization of K-means instead of random choosing. The results suggest that our proposed algorithm gets improved and consistent clusters for three real-world datasets than any other related system by providing 7.33\%, 7.90\%, and 0\% error in Iris, Letter, and Ruspini data. However, this system proved to be more useful in terms of computation time and CCPI when the number of clusters is more than 2.\\
Although our proposed system is much simpler and easier to implement than other existing systems mentioned, we have some limitations as well. Our method is highly prone to outliers. To implement this system, first outliers must be detected and eliminated if there are any.

\bibliographystyle{ieeetr}
\bibliography{main}

\end{document}